\documentclass[final,3p,times,twocolumn]{elsarticle}

\usepackage{amssymb}
\usepackage{amsthm}
\usepackage{amsmath}
\usepackage{graphicx}
\usepackage[caption=false]{subfig}

\usepackage{url}
\usepackage{booktabs}
\usepackage{color}
\usepackage{siunitx}
\usepackage{multirow}
\usepackage{lineno}
\usepackage{balance}

\begin{document}

\sloppy
\tolerance = 999

\newlength{\figw}
\newlength{\figh}

\begin{frontmatter}



\title{Improved Person Re-Identification Based on Saliency and Semantic Parsing \\ with Deep Neural Network Models}


\author[institute1]{Rodolfo Quispe}
\author[institute1]{Helio Pedrini}
\ead{helio@ic.unicamp.br}
\address[institute1]{Institute of Computing, University of Campinas, Campinas, SP, Brazil, 13083-852}

\begin{abstract}
Given a video or an image of a person acquired from a camera, person re-identification is the process of retrieving all instances of the same person from videos or images taken from a different camera with non-overlapping view. This task has applications in various fields, such as surveillance, forensics, robotics, multimedia. In this paper, we present a novel framework, named Saliency-Semantic Parsing Re-Identification (SSP-ReID), for taking advantage of the capabilities of both clues: saliency and semantic parsing maps, to guide a backbone convolutional neural network (CNN) to learn complementary representations that improves the results over the original backbones. The insight of fusing multiple clues is based on specific scenarios in which one response is better than another, thus favoring the combination of them to increase performance. Due to its definition, our framework can be easily applied to a wide variety of networks and, in contrast to other competitive methods, our training process follows simple and standard protocols. We present extensive evaluation of our approach through five backbones and three benchmarks. Experimental results demonstrate the effectiveness of our person re-identification framework. In addition, we combine our framework with re-ranking techniques to achieve state-of-the-art results on three benchmarks.
\end{abstract}


\end{frontmatter}


\section{Introduction}
\label{sec:intro}

Person re-identification (Re-ID) is a very challenging problem that aims to find all entities that have the same identification (ID) across cameras with respect to a gallery of individuals for a given probe (query). The probe and gallery are recorded from different camera views.

Some challenges associated with the Re-ID problem include occlusions, complex background, illumination conditions. However, the most difficult scenario is the occurrence of extreme changes in pose/viewpoint.

Re-ID is typically defined in a setting where no high resolution images are available (for example, security cameras installed in universities and airports). Since methods based on face recognition are not effective to be applied, current approaches are based on the appearance of people. More recently, the use of Convolutional Neural Networks (CNN) has become popular in this task.

All of the previously mentioned problems and constraints make Re-ID a difficult task, even for humans. Consider a scenario in which two different people are wearing similar clothing, with only a few difference in the colors of their belts and shoes. These details can be key clues to distinguishing people. In a security camera, small detail, such as the color of shoes or belt, may not be sufficiently clear so that they may be ignored by a human operator, where the two people would be considered as the same person. This realistic situation occurs in several Re-ID datasets, which present low resolution images, scaling changes, or misalignment in bounding boxes.

It is worth mentioning that the Re-ID task considers as input the bounding boxes around the people in the scene, which can be a sequence of images or videos. In this work, we focus on images, however, our framework can be easily extended to the video person Re-ID.

Since pose/viewpoint change are crucial issues for Re-ID, several approaches are based on separate horizontal-stripe images and compare people based on them, but such a method is not a complete solution. As pointed out by Kalayeh et al.~\cite{kalayeh2018human}, semantic parsing is a natural improvement for horizontal stripes because it provides labels at the pixel level, so we decided to use this insight in our framework. In addition, we realized that not every part of people is equally informative; in some cases, a backpack with a bright color or other salient objects may be a clue to the Re-ID. Thus, we designed a unified framework that unities semantic parsing and saliency to improve performance. The idea of combining multiple clues is natural to this problem, because each subnet stream of the framework can learn to solve different scenarios.

The main contributions of this paper are summarized as follows. Initially, we introduce a novel framework using saliency and semantic parsing. To the best of our knowledge, this is the first work that combines these two clues for Re-ID. Extensive experiments on three datasets and five backbones show the ability of our method to improve results and suggest that it can be used with many other backbones because of its definition. Different from other competitive methods, our framework takes full advantage of pretrained models and require a minimum number of fine-tuning epochs to reach competitive results. Moreover, our training process does not need to combine multiple Re-ID benchmarks.

Our framework, combined with re-ranking techniques, achieved state-of-the-art results on the three most widely used and challenging Re-ID datasets. We compared our work with the most competitive approaches available in the literature, yielding improvements of up to 4.1\% in mAP and 1.8\% rank-1.

The remainder of this paper is organized as follows. Section~\ref{sec:background} briefly reviews saliency detection, semantic parsing detection, as well as methods that use these concepts in the context of person re-identification. Section~\ref{sec:method} defines the re-identification problem and models that can be used as backbones of our framework, then it describes our method. Section~\ref{sec:results} offers implementation details, validation protocols, evaluation and comparison with the state-of-art. Section~\ref{sec:conclusions} concludes the paper with some final remarks and directions for future work.

\section{Background}
\label{sec:background}

This section reviews some relevant concepts and works associated with the research topic investigated in this work. Techniques for salient object detection, human semantic parsing, and person re-identification are described.

\subsection{Salient Object Detection}

Saliency detection is a task that aims to identify the fixation points that a human viewer would focus at the first glance. It has applications in various vision tasks, such as image segmentation, object detection, video summarization, compression, just to mention a few of them~\cite{wang2017deep}.

Early approaches to saliency detection were driven through local low-level features -- such as intensity, color, orientation and texture -- or global features based on finding regions in the image, which implies unique frequencies in the Fourier domain~\cite{goferman2012context}. In the last years, deep models have become the mainstream solution due to the CNNs capacity for representing multi-scale and multi-level features. Some current approaches include Multi-Layer Perceptrons (MLP) and Fully Convolutional Neural Network (FCNN)~\cite{borji2017salient}.

The first work that used the concept of saliency in the context of person re-identification was proposed by Zhao et al.~\cite{zhao2013person}. Their approach is based on a patch matching-based method. Each image patch has an associated saliency that is computed in an unsupervised fashion, then matching is computed inside the patch-neighborhood using hand-crafted features. A matching between patches with too different saliency brings a penalty to the model. Thus, the model is fitted to minimize the total cost of patch matching. Differently to this work, we do not use any patch matching-based approach and use deep features to encode person characteristics.

Liu et al.~\cite{liu2017hydraplus} proposed an attentive-based method, named HydraPlus-Net. Although the authors do not use the concept of saliency, the idea is related because they guide their network to focus more on specific regions of the image. Differently from this work, our approach first computes the salient object map from the input image and then uses this map to weigh an intermediate layer of the CNN backbone. Another difference is that our training pipeline does not include the saliency detection step as part of its process. Finally, our framework is designed to be capable to use different type of backbones (ResNet~\cite{he2016deep}, DenseNet~\cite{huang2017densely}, among others), whereas HydraPlus-Net is designed to use Inception~\cite{szegedy2015going} blocks for its construction.

Similar to HydraPlus-Net, Zhou et al.~\cite{zhou2018weighted} proposed to learn saliency maps and Re-ID at the same time. They introduced a weighted version of bilinear coding~\cite{lin2015bilinear} to encode higher-order channel-wise interactions. The main difference from our framework is that saliency maps are computed through the raw image in our pipeline, whereas the work by Zhou et al.~\cite{zhou2018weighted} uses the output of the GoogLeNet~\cite{szegedy2015going} as input to their saliency Part-Net.

Qian et al.~\cite{qian2017multi} proposed a network that learns saliency from their Re-ID pipeline. They accurately pointed out that features at different scales are not a well-solved problem for Re-ID. Their proposal, named MuDeep Net, is a network capable of learning features at different scales and creating saliency masks to emphasize channels with highly discriminative features. In our framework, we guide the network to learn from saliency and semantic parsing maps, without using multi-scale information.

\subsection{Human Semantic Parsing}

Human semantic parsing aims to segment human image into regions with fine-grained meaning, which has applications in Re-ID and human behavior analysis~\cite{Gong_2017_CVPR}. In its general form, semantic parsing has applications in several other domains, such as image montage, object colorization, stereo scene parsing, and medical segmentation~\cite{liu2015survey}.

Kalayeh et al.~\cite{kalayeh2018human} demonstrated that the use of semantic parsing can boost up results in Re-ID. They proposed to use an Inception-based network~\cite{szegedy2015going} that computes semantic maps and generates features for global representation of the input. Then, the feature map before the last average pooling is multiplied by the parsing maps to create a local representation. Our framework presents similarities with their method in the sense that we also employ human semantic parsing in Re-ID, but there are some key differences: the first one is that we use intermediate layers instead of the last one since we consider that very deep layer representation encodes too abstract information and it is not intuitive to combine it in a meaningful way with semantic and saliency maps. Second, we introduce saliency in our framework, as our experiments indicated, saliency and parsing maps contain complementary information able to enhance results. In fact, when integrated with re-ranking techniques, we achieved state-of-the-art performance. Finally, our training process does not require to combine various benchmarks for creating a huge training dataset.

\subsection{Person Re-Identification}

Person re-identification (Re-ID) is defined as the task of matching all instances of the same person across multiple cameras, that is, comparing a person of interest, named probe, against a gallery of candidates previously captured. Re-ID has applications related to surveillance of public areas/events for preventing dangerous events (such as terrorism and murder). For this reason, it has received attention from the computer vision community and has been widely studied over the last years.

Early works focused on handcrafted features such as color and texture, however, due to extreme viewpoint and illumination changes, these types of characteristics are not sufficiently discriminative. Currently, Deep Learning has established a new paradigm in the Re-ID problem.

Chang et al.~\cite{chang2018multi} proposed Multi-Level Factorization Net (MLFN) that encodes features at multiple semantic-levels. MLFN is composed of various stacked blocks with the same architecture and block selection modules that learn to interpret content of input images. The insight behind the selection blocks is to control and specialize the features that each block is learning.

Zhao et al.~\cite{zhao2017deeply} uses GoogLeNet~\cite{szegedy2015going} to extract features. Then, a multi-branch architecture uses these features to detect discriminative regions and create a part-aligned representation. From this idea, they were able to overcome misalignment and pose changes. Differently from this work, Su et al.~\cite{su2017pose} extracted person parts directly from the input images through a pose estimator trained independently. Then, they extracted features from complete images and parts. In the case of local clues, their architecture considers affine transformations. Finally, because pose estimation may be affected by pose changes or occlusions, they combined parts and global features using a weighted sub-net.

Li et al.~\cite{li2018harmonious} proposed Harmonious Attention Network (HA-CNN), which focuses on learning fine-grained relevant pixels and coarse latent regions at the same time. HA-CNN is based on Inception~\cite{szegedy2015going} blocks in a multi-branch structure for global and local representation. They further introduced a method for combining these representations in a harmonious way.

To compensate for viewpoint changes, Sarfraz et al.~\cite{sarfraz2018pose} proposed to create a pose-discriminative embedding. They trained a network that learns specialized features depending on the pose of the input: front, back or side. They also used body joint keypoints in order to guide the CNNs attention. Moreover, they proposed a new re-ranking technique, named Expanded Cross Neighborhood. Results were improved based on distance between gallery and probe features. Zhong et al.~\cite{zhong2017re} proposed a re-ranking method based on $k$-reciprocal nearest neighbors and Jaccard distance. It is worth mentioning that re-ranking methods are unsupervised and do not need any human interaction.

\section{Proposed Method}
\label{sec:method}

In this section, we describe the person re-identification problem more formally and present our framework.

\subsection{Problem Formulation}

We consider Re-ID as a retrieval process, that is, given a query person $x_p$ with ID $y_p$ and a gallery of $m$ people $X = \{ x_1, x_2, \ldots, x_m\}$ with IDs $Y = \{y_1, y_2, \ldots, y_m\}$, then Re-ID aims to recover all $x_i, (1 \leq i \leq m)$ such that $y_i = y_p$.

Suppose that a model $M(\theta)$ with learned parameters $\theta$ is capable of representing $x_p$ and people in $X$ with feature maps $f_p$ and $F = \{f_1, f_2, \ldots, f_m\}$, respectively. Thus, we can use Euclidean distance to compare $f_p$ against each element of $F$ and construct a ranked list based on the similarity of the feature maps. Depending on the application and context in which Re-ID is used, this ranked list may be cut off in the top 1, 5 or more. The resulting list $L$ (also referred as ranked list) is employed to represent only people that have identity equal to $y_p$\footnote{$L$ may not be totally correct, since $M(\theta)$ may not be perfect.}.

In this work, we create a model $M'$ that uses $M$ as backbone, such that the list $L'$ generated by $M'$ is \textit{better} than $L$. The quantitative definition of \textit{better} is based on the mean Average Precision (mAP) and cumulated matching characteristics (CMC). Both metrics are explained in the experiment section. Due to the definition of $M(\theta)$, our framework can be applied to many different backbones.

\subsection{Person Re-ID Framework}

\begin{figure}[!htb]
\centering
\setlength{\figw}{1.3cm}
\setlength{\figh}{2.7cm}
\includegraphics[width=\figw, height=\figh]{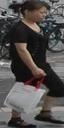} \hspace*{0.01cm}
\includegraphics[width=\figw, height=\figh]{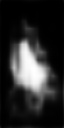} \hspace*{0.01cm}
\includegraphics[width=\figw, height=\figh]{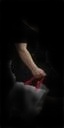} \\[0.2cm]
\includegraphics[width=\figw, height=\figh]{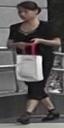} \hspace*{0.01cm}
\includegraphics[width=\figw, height=\figh]{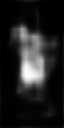} \hspace*{0.01cm}
\includegraphics[width=\figw, height=\figh]{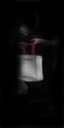}
\caption{Examples of saliency detection for the same person (from left to right): original image, saliency map, and result of overlap saliency map over the original image. The focus of the saliency is on the arm and white bag. Our framework uses this information to guide the feature learning process.}
\label{fig:salience-example}
\end{figure}

Based on the fact that a challenging issue for the re-identification task is caused by dramatic pose/viewpoint changes, we propose to combine global representation with saliency and semantic parsing masks. As shown in our experiments, these two types of masks generate complementary feature maps that improve results over the original CNN backbones. Saliency is important for Re-ID because in specific scenarios (Figure~\ref{fig:salience-example}) where people have certain items that can guide the re-identification process.

\begin{figure}[!htb]
\centering
\setlength\figw{1.1cm}
\setlength\figh{2.6cm}
\includegraphics[width=\figw, height=\figh]{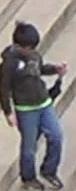} \hspace*{0.01cm}
\includegraphics[width=\figw, height=\figh]{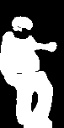} \hspace*{0.01cm}
\includegraphics[width=\figw, height=\figh]{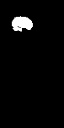} \hspace*{0.01cm}
\includegraphics[width=\figw, height=\figh]{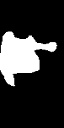} \hspace*{0.01cm}
\includegraphics[width=\figw, height=\figh]{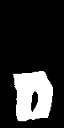} \hspace*{0.01cm}
\includegraphics[width=\figw, height=\figh]{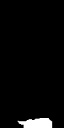} \\[0.2cm]
\includegraphics[width=\figw, height=\figh]{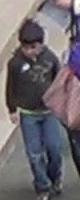} \hspace*{0.01cm}
\includegraphics[width=\figw, height=\figh]{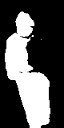} \hspace*{0.01cm}
\includegraphics[width=\figw, height=\figh]{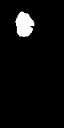} \hspace*{0.01cm}
\includegraphics[width=\figw, height=\figh]{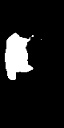} \hspace*{0.01cm}
\includegraphics[width=\figw, height=\figh]{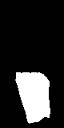} \hspace*{0.01cm}
\includegraphics[width=\figw, height=\figh]{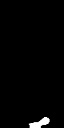}
\caption{Examples of parsing with five semantic regions of the same person with two different views. We use these maps to overcome misalignment and occlusions.}
\label{fig:parsing-example}
\end{figure}

However, saliency is not a complete solution to the problem because it focuses on some areas of the image and may be affected by occlusions. Thus, we use semantic parsing to encode every part of the person and overcome misalignment  in the bounding box detection and occlusions (Figure~\ref{fig:parsing-example}).

\begin{figure*}[!htb]
\centering
\includegraphics[width=16.5cm]{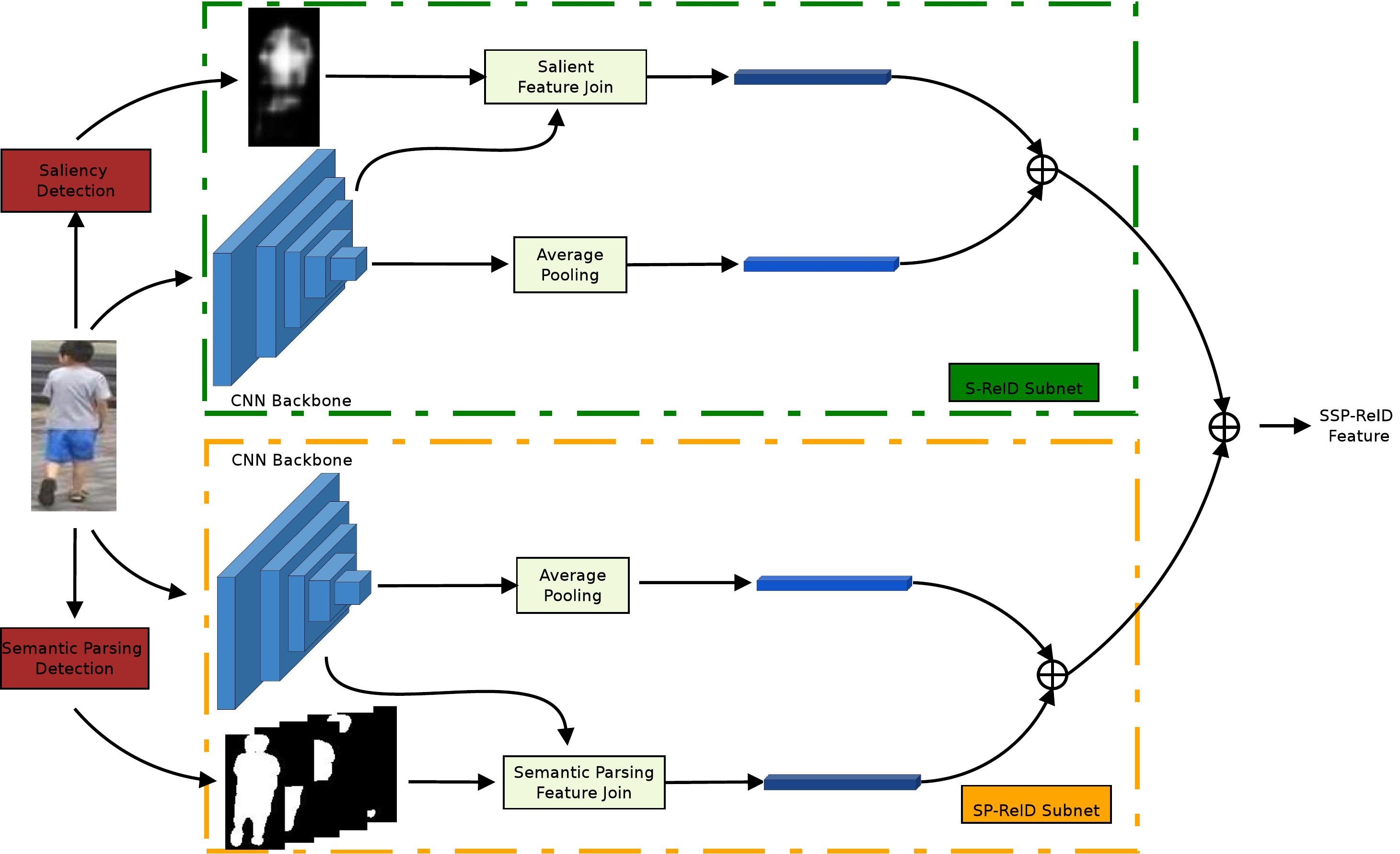}
\caption{SSP-ReID is a framework based on semantic parsing (SP-ReID subnet) and saliency (S-ReID subnet) to learn individual-similar performance representations. At the same time both representations are complementary because the union leads to an increase in performance.}
\label{fig:framework}
\end{figure*}

We propose the Saliency-Semantic Parsing (SSP-ReID) framework, as shown in Figure~\ref{fig:framework}, which is composed of two streams. Both of them have the same backbone architecture, however, without sharing weights. One of the streams (named S-ReID subnetwork) focuses on getting global-saliency features, whereas the other (named SP-ReID subnetwork) focuses on getting global-semantic-parsing features. The output of our framework is a feature map that is used to compare query and gallery images.

Given the input image, we compute the saliency and semantic parsing maps using off-the-shelf deep methods~\cite{li2016deepsaliency,Gong_2017_CVPR}. For the semantic parsing, we consider 5 semantic areas~\footnote{Head, upper body, lower body, shoes and complete body.}. Then, we use a CNN to create a global representation of the person. Moreover, we take the feature map from an intermediate layer and join it with saliency map in one stream, and semantic parsing maps in the other. We decide to use an intermediate layer since it is widely known that CNNs encode more abstract and higher semantic-level features as they go deeper (for instance, the relationship of head location between the input image and the very deep feature map may not be clear at first glance). Thus, it is more intuitive to combine raw saliency/semantic parsing maps with an intermediate layer as it does not have too abstract information and, at the same time, it encodes rich information.

Given an intermediate tensor $\tau \in \mathbb{R}^{h \times w \times c}$ and a saliency/semantic parsing map $\omega \in \mathbb{R}^{h' \times w'} $, in order to join intermediate feature tensor and saliency/semantic parsing information, we initially apply a bilinear interpolation over the tensor to transform $\tau \in \mathbb{R}^{h' \times w' \times c}$. Then, we apply element-wise product between every channel of the tensor and the map. Finally, we use average pooling to obtain the feature vector $v$. For the saliency feature join, the output feature is inside $\mathbb{R}^{c}$, whereas for semantic parsing feature join is inside $\mathbb{R}^{5c}$ due to the 5 semantic regions considered.

In order to train our network, we consider cross-entropy loss function with label smoothing regularizer (LSR)~\cite{szegedy2016rethinking} and triplet loss with hard positive-negative mining~\cite{hermans2017defense}.

Cross-entropy with LSR is defined as:
\begin{equation}
\begin{split}
H(q', p) & = - \sum_{k = 1}^{K} \log p(k)q'(k) \\
         & = (1 - \epsilon) H(q, p) + \epsilon H(u, p)
\end{split}
\end{equation}
\noindent where $K$ is the size of training batch, $\epsilon$ is a regularizer value, $p(k)$ is the output of the model, $q$ is the ground-truth distribution, $u$ is the uniform distribution and $q'$ is defined as :
\begin{equation}
q'(k) = (1 - \epsilon) q(k) + \frac{\epsilon}{K}
\end{equation}

LSR is a change in the ground-truth labels distribution, which aims to make the model more adaptable by adding prior distribution over the labels. We consider this loss over general cross entropy in order to avoid the largest logit from becoming much larger than all others, this prevents overfit.

Triplet loss with hard positive-negative mining is defined as:
\begin{equation}
    \begin{split}
    T(X) = \sum_{i = 1}^{P} \sum_{a = 1}^{N} [ m + \max_{p = 1 \ldots N}D(f(x_a^i), f(x_p^i)) \\
    - \min_{\substack{p = 1 \ldots N \\ n = 1 \ldots N \\ i \not = j}}D(f(x_a^i), f(x_n^j))]_{+}
    \end{split}
\end{equation}
\noindent where $X$ is a training batch, with $P$ people and $N$ images per person, $f(~.~)$ is the output feature map of the network, $x_j^i$ is the $j$-th image of the $i$-th person, $D(~.~,~.~)$ is a distance function (e.g. Euclidean), $[\sigma]_+$ denotes $\max(\sigma~,~0)$ and $m$ is hyperparameter named margin. Basically, this loss finds the pair of images of the same person with maximum distance and the pair of images of different people with minimum distance and guides the model to make the difference between these two at least equal to the margin $m$.

\begin{figure*}[!htb]
\centering
\setlength{\figw}{16.5cm}
\includegraphics[width=\figw]{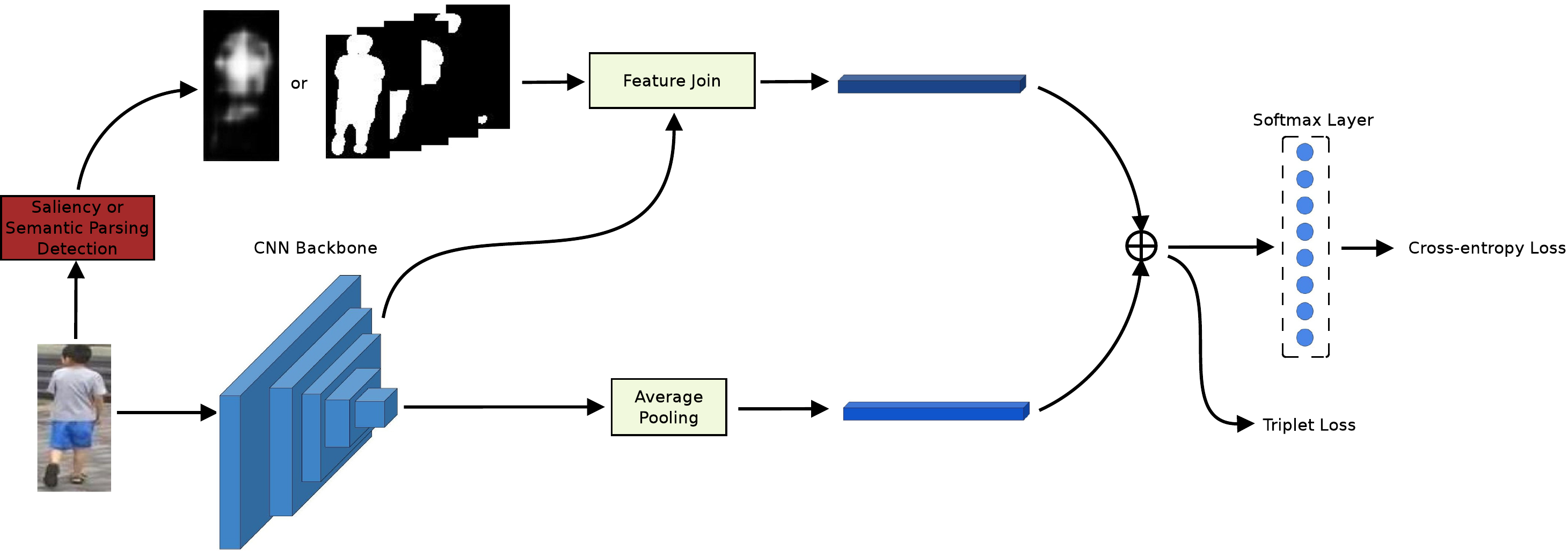}
\caption{Training setup for S-ReID and SP-ReID subnetworks. When training out framework, we consider triplet and cross-entropy loss functions. For the triplet loss, we take the feature vector before softmax layer and use it to compare images based on the Euclidean distance. The triplet loss may be ignored depending on the CNN backbone.}
\label{fig:framework-train}
\end{figure*}

SP-ReID and S-ReID subnetworks are trained separately and, depending on the backbone, we use the sum of both losses or only cross-entropy with LSR. To train our network, we add a multi-class classification layer to the end of the subnetwork. Figure~\ref{fig:framework-train} illustrates the network architecture for the training step, as well as its relation to the loss functions.

\section{Experimental Results}
\label{sec:results}

In this section, the parameter setting and datasets used in our experiments will be first introduced. Next, we will show the results obtained with our proposed method and compare them with state-of-the-art approaches.

\subsection{Implementation Details}

The saliency detection is performed via the off-the-shelf FCNN proposed by Li et al.~\cite{li2016deepsaliency}, whereas the semantic parsing detection is computed through the Joint Human Parsing and Pose Estimation Network (JJPNet)~\cite{Gong_2017_CVPR} trained in the Look into Person (LIP) dataset~\cite{liang2018look}.

We evaluate our framework using 5 different backbones. Table~\ref{table:backbones-loss-intermediate-layer} summarizes the loss function used for each backbone. The initial weights of all backbones are Imaginet~\cite{russakovsky2015imagenet} pretrained models. In the case of the intermediate layer to be combined with saliency and semantic parsing maps: (i) we use the output of layer Res5C for ResNet50~\cite{he2016deep} and ResNet50-M~\cite{yu2017devil}; (ii) we use the output last Inception-B block for Inception-V4~\cite{szegedy2017inception}; (iii) we use the output of Middle Flow layer for Xception~\cite{chollet2016xception}; and (iv) we use the output of the second composite function for DenseNet121~\cite{huang2017densely}.

\begin{table}[!htb]
\centering
\renewcommand{\tabcolsep}{2.5mm}
\renewcommand{\arraystretch}{1.1}
\caption{Loss function used for each backbone. CROSSE stands for Cross Entropy with LSR~\cite{szegedy2016rethinking}, whereas TRIP stands for Triplet Loss with hard positive-negative mining~\cite{hermans2017defense}. Using TRIP in Inception-V4~\cite{szegedy2017inception} and Xception~\cite{chollet2016xception} raises exploding gradient.}
\label{table:backbones-loss-intermediate-layer}
\small
\begin{tabular}{lc}
\toprule
\textbf{Backbone} &\textbf{Loss function} \\
\midrule
ResNet50~\cite{he2016deep}               & TRIP + CROSSE \\
Densenet121~\cite{huang2017densely}      & TRIP + CROSSE \\
Resnet50-M~\cite{yu2017devil}            & TRIP + CROSSE \\
Inception-V4~\cite{szegedy2017inception} & CROSSE \\
Xception~\cite{chollet2016xception}      & CROSSE \\
\bottomrule
\end{tabular}
\end{table}

We adjust the size of the input to $254 \times 128$ pixels and saliency/semantic parsing maps to $128 \times 64$ pixels. Adam optimizer is used with training batch of $32$, initial learning rate of $0.0003$, weight decay of $0.0005$, and a learning rate decay factor of $0.1$ every $60$ epochs. We also fix the number of training epochs to $180$ for every backbone. In the LSR implementation, we set $\epsilon = 0.1$ and triplet loss as $m = 0.3$. Finally, we use the re-ranking method proposed by Zhong et al.~\cite{zhong2017re} to compare our results with state-of-the-art approaches~\footnote{Code and models will be available upon acceptance.}.

\subsection{Datasets and Validation Protocols}

We evaluate our framework on three widely used datasets. A summary of them is shown in Table~\ref{table:datasets}, which reports the number of people, bounding boxes and cameras present in each benchmark setup.

\begin{table}[!htb]
\small
\setlength{\tabcolsep}{1.4mm}
\renewcommand{\arraystretch}{1.1}
\centering
\caption{Comparative summary of Re-ID datasets used in our experiments.}
\begin{tabular}{lccc}
\toprule
\textbf{Dataset} & \# \textbf{People} & \# \textbf{BBox} & \# \textbf{Cameras} \\
\midrule
Market1501~\cite{zheng2015scalable}     & 1501 & 32668 & 6 \\
CUHK03~\cite{li2014deepreid}            & 1467 & 14096 & 6 \\
DukeMTMC-ReID~\cite{zheng2017unlabeled} & 1812 & 36411 & 8 \\
\bottomrule
\end{tabular}
\label{table:datasets}
\end{table}

The DukeMTMC-ReID dataset~\cite{zheng2017unlabeled} is a subset of the DukeMTMC dataset~\cite{ristani2016MTMC} for image-based re-identification with hand-drawn bounding boxes. Bounding boxes of different sizes with outdoor scenes as background are available. For validation, we use the fixed training and testing sets proposed in the original protocol of the dataset.

The Market1501~\cite{zheng2015scalable} was created through Deformable Part Model (DPM) in order to simulate a real-world scenario. For validation, we use the fixed training and testing sets provided with the dataset.

The CUHK03~\cite{li2014deepreid} has an average of 4.8 images per view. Misalignment, occlusions and missing body parts are quite common. For validation, we use the new validation protocol~\cite{zhong2017re} with partition of 767/700. Moreover, we evaluate detected (CUHK03 (D)) and labeled (CUHK03 (L)) versions of the dataset.

Quantitative results for every dataset are based on mean Average Precision (mAP) and Cumulative Matching Curve (CMC). The mAP considers the order in which the gallery is sorted for a given query, defined as:

\begin{equation}
mAP = \frac{AP}{\# queries}
\end{equation}

\noindent where $AP$ is defined as

\begin{equation}
AP = \normalsize \frac{\displaystyle \sum_{k=1}^n P(k)\cdotp rel(k)}{\# relevant~items}
\end{equation}

\noindent where $n$ is the number of recovered items, $rel(k)$ is equal to 1 if the $k$-th item is relevant to the query and 0 otherwise, and $P(k)$ is defined as:

\begin{equation}
P(k) = \normalsize \frac{\displaystyle\sum_{i=1}^k \displaystyle\normalsize rel(i)}{k}
\end{equation}

The CMC represents the probability that a correct match with the query identity will appear in variable-sized ranked list:

\begin{equation}
CMC(r) = \frac{in(r)}{\# queries}
\end{equation}

\noindent where $in(r)$ is the number of queries that have a relevant element within the first $r$ items in the ranked list. We set $r=1$ and refer to it as rank-1.

\subsection{Performance in Re-ID}

In this section, we evaluate and compare different aspects of our framework: backbone (e.g., ResNet~\cite{he2016deep}), saliency subnet (S-ReID), semantic parsing subnet (SP-ReID) and the complete framework (SSP-ReID). Results are summarized in Table~\ref{table:results-re-id}. Overall, ResNet50-M + SSP-ReID produced the best results for all datasets, whereas there are marginal differences in using ResNet50~\cite{he2016deep} and DenseNet~\cite{huang2017densely} as backbones. On the other hand, Inception-V4~\cite{szegedy2017inception} and Xception~\cite{chollet2016xception} yielded the worse performance. Note that DenseNet~\cite{huang2017densely} gets interesting results despite of its lower number of parameters. In addition, our framework raises consistently an improvement over all backbones, this suggest that our framework can be used as a enhancing method in future works.

\begin{table*}[!htb]
\renewcommand{\arraystretch}{1.1}
\setlength{\tabcolsep}{1.1mm}
\centering
\caption{Results of framework in Re-ID. ResNet + S-Reid stands for Saliency subnet using ResNet as backbone. Analogously, SP-ReID refers to Semantic Parsing subnet, whereas SSP-ReID refers to the complete framework. We highlight in red color the cases in which the subnetwork/framework is worse than the original backbone, whereas cases with better results than backbone are highlighted in blue color.}
\label{table:results-re-id}
\small
\begin{tabular}{lcccccccc}
\toprule
&\multicolumn{2}{c}{\textbf{Market1501}} &\multicolumn{2}{c}{\textbf{CUHK03 (D)}} &\multicolumn{2}{c}{\textbf{CUHK03 (L)}} & \multicolumn{2}{c}{\textbf{DukeMTMC-reID}} \\
\midrule
\textbf{Method} & \textbf{mAP(\%)} & \textbf{rank-1(\%)} & \textbf{mAP(\%)} & \textbf{rank-1(\%)} & \textbf{mAP(\%)} & \textbf{rank-1(\%)} & \textbf{mAP(\%)} & \textbf{rank-1(\%)} \\
\midrule
ResNet~\cite{he2016deep} & 72.9 & 88.1 & 52.9 & 55.6 & 56.7 & 58.8 & 62.1 & 77.7 \\
ResNet + S-ReID     & \textcolor{blue}{73.0} & \textcolor{red}{87.6} & \textcolor{blue}{53.4} & \textcolor{red}{56.0} & \textcolor{red}{54.4}    & \textcolor{red}{55.9} & \textcolor{blue}{63.1} & \textcolor{blue}{78.9} \\
ResNet + SP-ReID    & \textcolor{red}{72.4} & \textcolor{red}{87.8} & \textcolor{blue}{53.5} & \textcolor{red}{56.2} & \textcolor{red}{55.5}     & \textcolor{red}{57.3} & \textcolor{blue}{62.7} & \textcolor{blue}{78.0} \\
ResNet + SSP-ReID   & \textcolor{blue}{75.9} & \textcolor{blue}{89.3} & \textcolor{blue}{57.1} & \textcolor{blue}{59.4} & \textcolor{blue}{58.9} & \textcolor{blue}{60.6} & \textcolor{blue}{66.1} & \textcolor{blue}{80.1} \\
\midrule
ResNet-M~\cite{yu2017devil}         & 77.5 & 91.2 & 56.3 & 58.7 & 58.9 & 61.1 & 63.5 & 78.8 \\ 
ResNet-M + S-ReID   & \textcolor{blue}{77.6} & \textcolor{blue}{91.2} & \textcolor{blue}{56.7} & \textcolor{blue}{59.4} & \textcolor{blue}{59.7} & \textcolor{blue}{62.1} & \textcolor{blue}{65.2} & \textcolor{blue}{80.6} \\
ResNet-M + SP-ReID  & \textcolor{red}{76.6} & \textcolor{red}{90.9} & \textcolor{blue}{57.3} & \textcolor{blue}{59.9} & \textcolor{blue}{59.7}   & \textcolor{blue}{61.4} & \textcolor{blue}{64.9} & \textcolor{blue}{79.6} \\
ResNet-M + SSP-ReID & \textcolor{blue}{80.1} & \textcolor{blue}{92.5} & \textcolor{blue}{60.5} & \textcolor{blue}{63.1} & \textcolor{blue}{63.3} & \textcolor{blue}{65.6} & \textcolor{blue}{68.6} & \textcolor{blue}{81.8} \\
\midrule
DenseNet~\cite{huang2017densely} & 72.0 & 89.3 & 42.2& 44.1 & 45.6 & 47.4 & 62.5 & 79.7 \\
DenseNet + S-ReID    & \textcolor{blue}{72.3} & \textcolor{blue}{89.7} & \textcolor{blue}{43.1} & \textcolor{blue}{44.9} & \textcolor{red}{44.3} & \textcolor{red}{46.7} & \textcolor{blue}{62.6} & \textcolor{blue}{80.3} \\
DenseNet + SP-ReID   & \textcolor{blue}{72.9} & \textcolor{blue}{89.6} & \textcolor{blue}{43.3} & \textcolor{blue}{44.6} & \textcolor{red}{44.3} & \textcolor{red}{44.9} & \textcolor{blue}{62.9} & \textcolor{blue}{79.8} \\
DenseNet + SSP-ReID  & \textcolor{blue}{76.7} & \textcolor{blue}{90.9} & \textcolor{blue}{48.1} & \textcolor{blue}{48.1} & \textcolor{blue}{49.5} & \textcolor{blue}{49.1} & \textcolor{blue}{67.1} & \textcolor{blue}{82.2} \\
\midrule
Inception-V4~\cite{szegedy2017inception} & 64.0 & 81.9 & 38.7 & 38.7 & 40.7 & 42.4 & 49.6 & 71.9 \\
Inception-V4 + S-ReID     & \textcolor{red}{62.8} & \textcolor{red}{81.4} & \textcolor{blue}{41.2} & \textcolor{blue}{43.1} & \textcolor{blue}{42.5} & \textcolor{blue}{42.4} & \textcolor{red}{49.1} & \textcolor{red}{70.6} \\
Inception-V4 + SP-ReID    & \textcolor{red}{62.1} & \textcolor{red}{80.6} & \textcolor{red}{34.7} & \textcolor{red}{35.6} & \textcolor{red}{36.1} & \textcolor{red}{37.4} & \textcolor{red}{49.0} & \textcolor{red}{70.6} \\
Inception-V4 + SSP-ReID   & \textcolor{blue}{67.7} & \textcolor{blue}{85.4} & \textcolor{blue}{45.5} & \textcolor{blue}{46.4} & \textcolor{blue}{45.5} & \textcolor{blue}{45.2} & \textcolor{blue}{55.0} & \textcolor{blue}{75.5} \\
\midrule
Xception~\cite{chollet2016xception} & 50.1 & 69.9 & 26.0 & 26.3 & 25.2 & 25.2 & 36.1 & 55.4 \\
Xception + S-ReID     & \textcolor{red}{49.8} & \textcolor{red}{68.2} & \textcolor{red}{23.7} & \textcolor{red}{23.4} & \textcolor{red}{24.2} & \textcolor{red}{24.1} & \textcolor{red}{33.6} & \textcolor{red}{52.9} \\
Xception + SP-ReID    & \textcolor{red}{47.5} & \textcolor{blue}{70.9} & \textcolor{red}{20.7} & \textcolor{red}{22.9} & \textcolor{red}{20.8} & \textcolor{red}{21.6} & \textcolor{red}{34.6} & \textcolor{blue}{56.6} \\
Xception + SSP-ReID   & \textcolor{blue}{57.5} & \textcolor{blue}{77.1} & \textcolor{blue}{29.4} & \textcolor{blue}{29.9} & \textcolor{blue}{30.0} & \textcolor{blue}{29.6} & \textcolor{blue}{42.8} & \textcolor{blue}{63.9} \\
\bottomrule
\end{tabular}
\end{table*}

\begin{figure}[!htb]
\setlength{\figw}{5.0cm}
\centering
\includegraphics[width=\figw]{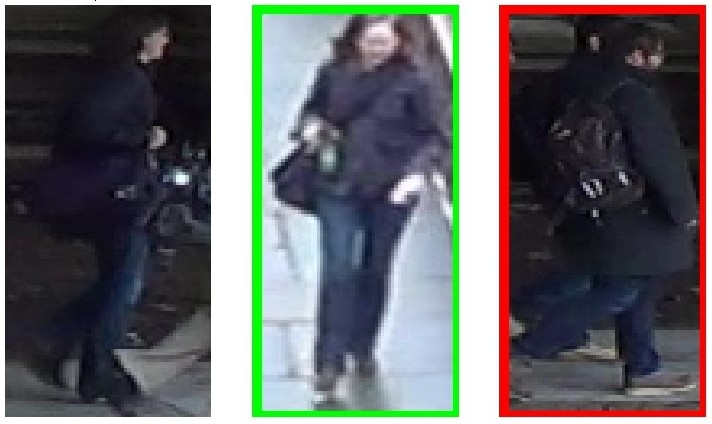} \hspace*{0.01cm}
\includegraphics[width=\figw]{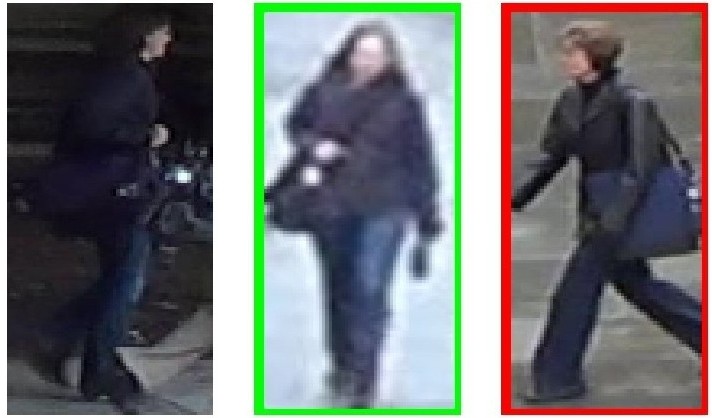} \hspace*{0.01cm}
\includegraphics[width=\figw]{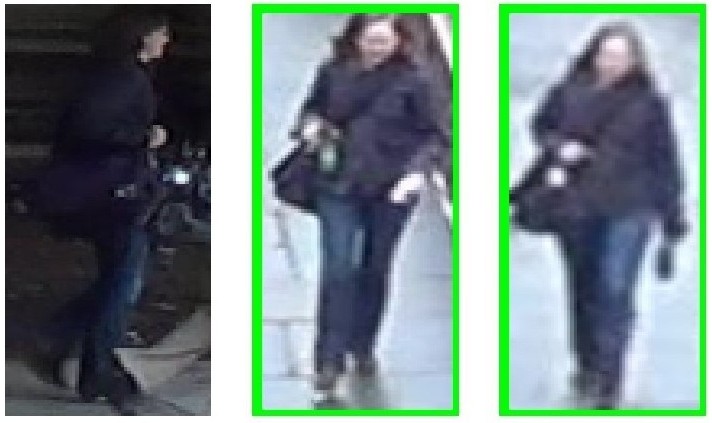}
\caption{Qualitative example of performance of framework modules. The first column represents the query, the second and third columns represent the results of rank-1 and rank-2, respectively (the correct results are in green, otherwise they are in red). The first, second and third rows represent the output of S-ReID, SP-ReID and SSP-ReID, respectively. For the same query, S-ReID and SP-ReID achieved different correct rank-1 results and wrong rank-2 results. On the other hand, SSP-ReID can combine the best characteristics of both subnets and reach correct rank-2.}
\label{fig:sub-net-example}
\end{figure}

In the light of all datasets, S-ReID achieved improvements up to 1.3\% for mAP and up to 4.4\% for rank-1 over individual backbones, however, in general the improvements were marginal. There are also cases where results were marginally worse. This same scenario is repeated for SP-ReID, but if we consider the complete framework (combination of S-ReID and SP-ReID), we consistently obtained better results, with improvements up to 7.4\% (mAP) and 7.2\% (rank-1) in Market1501, 6.8\%(mAP) and 7.7\%(rank-1) in CUHK03 (D), 4.9\%(mAP) and 4.5\%(rank-1) in CUHK03 (L), 6.7\%(mAP) and 8.5\% (rank-1) for DukeMTMC-reID. This suggests that S-ReID and SP-ReID are learning similar performance representation, but with complementary information, which improves the model capacity when they are combined. Figure~\ref{fig:sub-net-example} shows an example of rank-2 results for DukeMTMC-ReID~\cite{zheng2017unlabeled}.

It can be observed that the method improvement is inversely proportional to the capacity of the backbone. For better backbones (for instance, ResNet50-M~\cite{yu2017devil}), the improvements are smaller when compared to the lower performance backbones (for instance, Xception~\cite{chollet2016xception}). This is related to the complexity of datasets: as we start achieving high results in mAP or rank-1, we need much more higher discriminative models that can deal with more specific and often sparse cases.

\subsection{Performance Comparison}

We evaluate our method with various competitive approaches available in the literature. A performance comparison is presented in Table~\ref{table:comparison-state-of-art}, where we applied re-ranking~(RR) to boost our final results.

\begin{table*}[!htb]
\renewcommand{\arraystretch}{1.1}
\setlength{\tabcolsep}{1.1mm}
\centering
\caption{Comparison with the state-of-art, in \textbf{bold} the best results, RR stands for re-ranking}
\label{table:comparison-state-of-art}
\small
\begin{tabular}{lcccccccc}
\toprule
&\multicolumn{2}{c}{\textbf{Market1501}} &\multicolumn{2}{c}{\textbf{CUHK03 (D)}} &\multicolumn{2}{c}{\textbf{CUHK03 (L)}} & \multicolumn{2}{c}{\textbf{DukeMTMC-reID}} \\
\midrule
\textbf{Method} & \textbf{mAP(\%)} & \textbf{rank-1(\%)} & \textbf{mAP(\%)} & \textbf{rank-1(\%)} & \textbf{mAP(\%)} & \textbf{rank-1(\%)} & \textbf{mAP(\%)} & \textbf{rank-1(\%)} \\
\midrule
SPreID~\cite{kalayeh2018human}         & 83.3 & 93.6 & ---  & ---  & ---  & ---  & 73.3 & 85.9 \\
DaRe(De)+RE+RR~\cite{wang2018resource} & 86.7 & 90.9 & 71.6 & 70.6 & 74.7 & 73.8 & 80.0 & 84.4 \\
DuATM~\cite{si2018dual}                & 76.6 & 91.4 & ---  & ---  & ---  & ---  & 64.5 & 81.8 \\
HA-CNN~\cite{li2018harmonious}         & 75.7 & 91.2 & 41.0 & 44.4 & 38.6 & 41.7 & 63.8 & 80.5 \\
ATWL~\cite{ristani2018features}        & 75.6 & 89.4 & ---  & ---  & ---  & ---  & 63.4 & 79.8 \\
PSE~\cite{sarfraz2018pose}             & 84.0 & 90.3 & ---  & ---  & ---  & ---  & 79.8 & 85.2 \\
MLFN~\cite{chang2018multi}             & 74.3 & 90.0 & 47.8 & 52.8 & 49.2 & 54.7 & 62.8 & 81.0 \\
SVDNet~\cite{sun2017svdnet}            & 62.1 & 82.3 & 37.8 & 40.9 & 37.2 & 41.5 & 56.8 & 76.7 \\
\midrule
ResNet + SSP-ReID                      & 75.9 & 89.3 & 57.1 & 59.4 & 58.9 & 60.6 & 66.1 & 80.1 \\
ResNet + SSP-ReID + RR                 & 88.2 & 91.5 & 71.1 & 67.6 & 72.4 & 68.4 & 81.4 & 84.8 \\
ResNet-M + SSP-ReID                    & 80.1 & 92.5 & 60.5 & 63.1 & 63.3 & 65.6 & 68.6 & 81.8 \\
ResNet-M + SSP-ReID + RR               & \textbf{90.8} & \textbf{93.7} & \textbf{75.0} & \textbf{72.4} & \textbf{77.5} & \textbf{74.6} & \textbf{83.7} & \textbf{86.4} \\
DenseNet + SSP-ReID                    & 76.7 & 90.9 & 48.1 & 48.1 & 49.5 & 49.1 & 67.1 & 82.2 \\
DenseNet + SSP-ReID + RR               & 89.9 & 93.3 & 63.1 & 58.4 & 64.7 & 59.9 & 83.3 & 86.2 \\
\bottomrule
\end{tabular}
\end{table*}

Our method was able to achieve state-of-the-art in all datasets. Overall, performance improvement is higher for mAP than rank-1, because mAP considers a greater number of elements in its definition. Thus, it is more sensitive to any change in the ranking list. SPreID~\cite{kalayeh2018human} is the method with the closest performance, however, unlike such approach, our framework does not have to be trained with 10 datasets, we only need 180 epochs in each dataset. In addition, our framework is easier to implement compared to other methods~\cite{chang2018multi,li2018harmonious,wang2018resource,ristani2018features}.

\section{Conclusions and Future Work}
\label{sec:conclusions}

In this work, we presented SSP-ReID, a framework based on saliency and semantic parsing information, which achieved state-of-the-art results on three challenging datasets for the re-identification task. SSP-ReID is composed of two subnetworks, a saliency-guided subnet that aims to focus on learning in specific parts of the image and a semantic parsing-guided subnet for dealing with misalignments, occlusions, and other challenging issues for the person re-identification task.

We conducted extensive evaluation of our framework on five different backbones and three datasets. The representation learned from the saliency-guided and semantic parsing-guided subnetworks has similar performance to that of the individual backbones, however, both combined boost the results, indicating that the learned representation is complementary.

Our framework can be easily adapted to multiple CNN backbones, further improving performance over original networks. We expect that the use of multiple clues (for instance, human pose and multi-scale strategies) inspires other re-identification works.

\section{Acknowledgments}
\label{acknowledgment}

The authors are thankful to FAPESP (grant \#2014/12236-1) and CNPq (grant \#305169/2015-7) for their financial support.



\bibliographystyle{elsarticle-num}
\bibliography{ms}

\end{document}